%% file: main.tex
\documentclass[10pt,twocolumn,letterpaper]{article}

\usepackage[pagenumbers]{cvpr} 
\usepackage[accsupp]{axessibility} 
\usepackage{multirow}
\usepackage{siunitx}
\usepackage{caption}
\usepackage{subcaption}
\usepackage{algpseudocode}
\input{preamble}

\definecolor{cvprblue}{rgb}{0.21,0.49,0.74}
\usepackage[pagebackref,breaklinks,colorlinks,citecolor=cvprblue]{hyperref}
\usepackage{soul}

\newcommand{\mbf}[1]{\ensuremath{\mathbf{#1}}}

\def\paperID{189} 
\def\confName{CVPR}
\def\confYear{2024}

\title{Event-based Structure-from-Orbit}

\author{Ethan Elms$^\dagger$ \quad Yasir Latif$^\dagger$ \quad Tae Ha Park$^\ddagger$ \quad Tat-Jun Chin$^\dagger$%
\thanks{SmartSat CRC Professorial Chair of Sentient Satellites.}\\
$^\dagger$The University of Adelaide \hspace{20px} $^\ddagger$Stanford University\\
{\tt\small {\{ethan.elms,yasir.latif,tat-jun.chin\}}@adelaide.edu.au$^\dagger$},%
{\tt\small {tpark94}@stanford.edu$^\ddagger$}%
}

\begin{document}
\maketitle

\begin{abstract}
Event sensors offer high temporal resolution visual sensing, which makes them ideal for perceiving fast visual phenomena without suffering from motion blur. Certain applications in robotics and vision-based navigation require 3D perception of an object undergoing circular or spinning motion in front of a static camera, such as recovering the angular velocity and shape of the object. The setting is equivalent to observing a static object with an orbiting camera. In this paper, we propose event-based structure-from-orbit (eSfO), where the aim is to simultaneously reconstruct the 3D structure of a fast spinning object observed from a static event camera, and recover the equivalent orbital motion of the camera. Our contributions are threefold: since state-of-the-art event feature trackers cannot handle periodic self-occlusion due to the spinning motion, we develop a novel event feature tracker based on spatio-temporal clustering and data association that can better track the helical trajectories of valid features in the event data. The feature tracks are then fed to our novel factor graph-based structure-from-orbit back-end that calculates the orbital motion parameters (\eg, spin rate, relative rotational axis) that minimize the reprojection error. For evaluation, we produce a new event dataset of objects under spinning motion. Comparisons against ground truth indicate the efficacy of eSfO.
\end{abstract}

\section{Introduction}\label{sec:intro}

Three-dimensional (3D) perception is a fundamental vision capability~\cite{COLMAP,ozyecsil2017survey}. Recent works have focused on the use of neuromorphic event sensors \cite{gallego2020event} for
 3D perception tasks, such as visual odometry (VO)~\cite{zhou2021event,kueng2016low}, structure-from-motion (SfM) and simultaneous localization and mapping (SLAM)~\cite{vidal2018ultimate,mueggler2017event}. 
Event sensors offer several advantages over conventional cameras, such as high temporal resolution, low power and low data rate asynchronous sensing. Event sensors also offer a higher dynamic range, enabling them to see more details in difficult lighting conditions.

\begin{figure}[t]\centering
\includegraphics[width=0.5\textwidth]{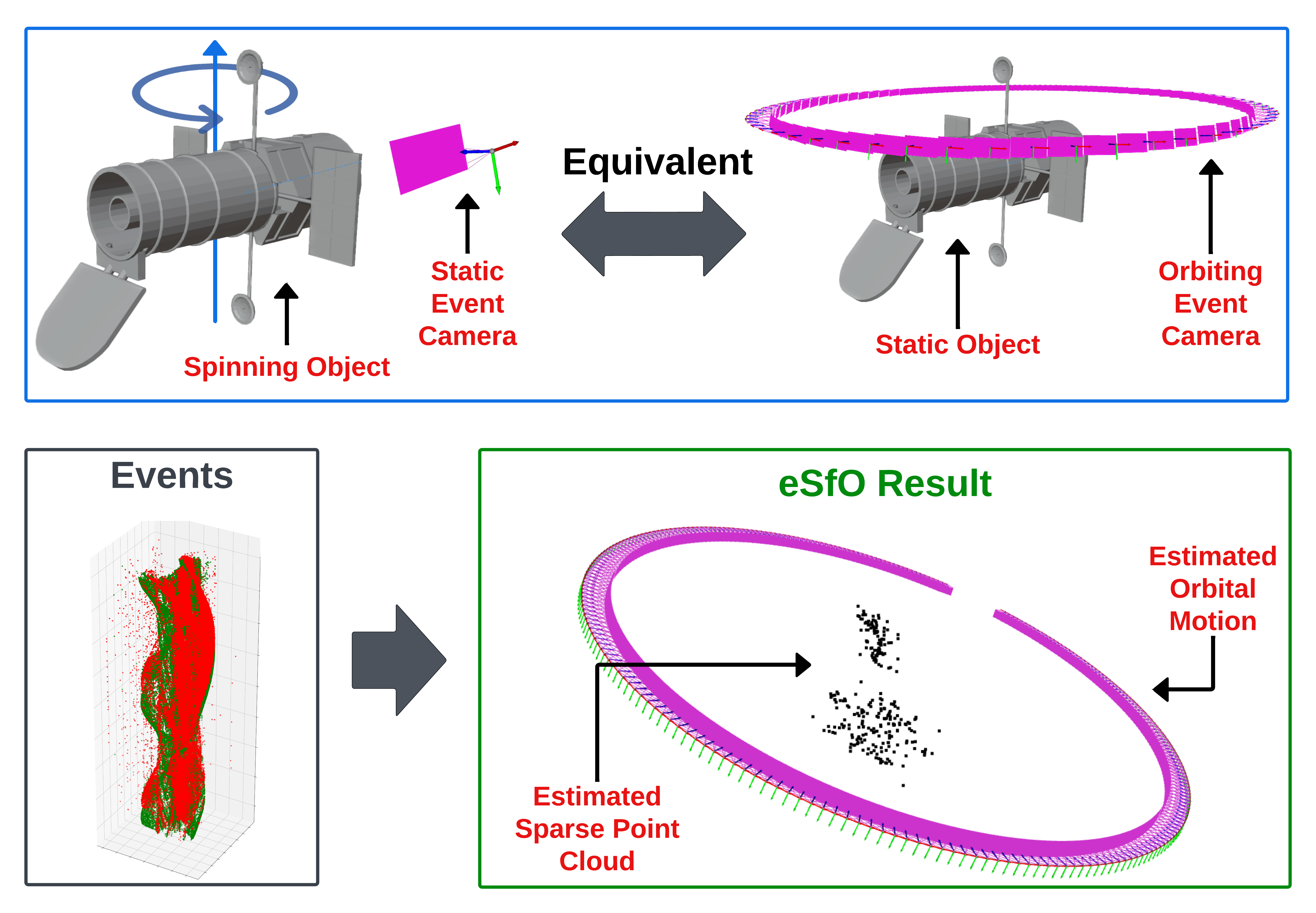}
    \caption{In eSfO, we exploit the equivalence between a static event camera observing a spinning object, and an orbiting event camera observing a static object. This enables us to jointly estimate the motion parameters (\eg, spin rate, rotational axis relative to the camera), as well as the sparse structure of the object.}
    \label{fig:teaser}
\end{figure}

However, the advantages of event sensors come with certain limitations. Firstly, due to the asynchronous nature of events, the idea of a consistent local neighborhood for a pixel across multiple views no longer holds. Moreover, image gradients -- a fundamental invariant in intensity images often used as basis for feature detection, description, and tracking -- are not observable in event data. This makes feature detection and tracking challenging, which represents a major obstacle towards event-only SLAM systems. 

In this work, we formulate the task of \emph{event-based Structure-from-Orbit (eSfO)}, which entails the reconstruction of an object undergoing circular motion while being observed by a static event camera, and jointly estimating the motion of the object; see Fig.~\ref{fig:teaser}. An object is undergoing a circular motion if it is rotating about a fixed axis, \emph{a.k.a.}~spinning. Since observing a spinning object from a static camera is mathematically equivalent to  observing a static object from an orbiting camera, the motion of the spinning object can be recovered as the motion of the orbiting camera. Using an event sensor to observe the spinning target is compelling, since the effect of motion blur is alleviated, especially in cases with high angular velocity.

SfM of a spinning object, usually placed on a turntable, has been studied previously~\cite{fitzgibbon1998automatic,fremont2004turntable}. However, these early works assume known angular rates and solve the task of structure recovery only. Nonetheless, various e-commerce, multimedia and augmented reality applications can benefit from such a 3D reconstruction approach~\cite{niem1999automatic,park2003stereo,park2005multiview}.

Outside of turntable-induced rotations, spinning objects can be observed naturally in the world, such as race-cars spinning in place \cite{cheng2023accidental} allowing opportunistic observation of the complete structure. Space is another domain abundant with naturally spinning objects, from as large as celestial bodies such as planets and asteroids to as small as satellites and space debris. For instance, as most small bodies such as asteroids tend to spin about the major principal axis, many works leveraged the single-axis spin assumption for 3D reconstruction purposes \cite{bandyonadhyay_2019_aero_sfs, dor_2021_cvprw_visual, dennison_2023_taes_ans}. Such motion characteristics are also observed on some man-made objects freely rotating in inertial space \cite{tweddle_2015_jfr_slam}. Indeed, \citet{kaplan_2010_aiaa_debris} reported in 2010 that there are over 100 expired satellites in geosynchronous orbits (GEO) spinning at high angular rates (tens of RPM) as they retain the angular momentum from spin-stabilization during service. Extracting the shape and motion parameters of such spinning objects in space can facilitate vision-based spacecraft navigation~\cite{dor_2021_cvprw_visual,dennison_2023_taes_ans} and formation flying~\cite{guffanti2023autonomous}.

\paragraph{Contributions}
Monocular eSfO is challenging since current event-based feature detection and tracking methods are not reliable on spinning objects that periodically self-occlude. Moreover, the underlying structure-from-orbit (SfO) problem imposes more constraints than SfM, and SfO has not been satisfactorily tackled in the literature. Our work addresses the difficulties by contributing:
\begin{itemize}
\item A novel eSfO formulation that takes into account the problem structure induced by a spinning object (Sec.~\ref{sec:SfS}).
\item A novel event feature detection and tracking mechanism designed for the spinning object case (Sec. \ref{sec:tracking}).
\item A novel factor graph-based optimization back-end that efficiently estimates sparse structure and orbit motion parameters from event feature tracks (Sec.~\ref{sec:esfo_details}).
\item A monocular event sensor dataset and benchmark for the eSfO problem (Sec. \ref{sec:dataset}).
\end{itemize}

\section{Related Work}

\subsection{Event-based Vision}
Event sensors have been incorporated into various SLAM and VO pipelines, due to their low data rate and high dynamic range~\cite{huang2024eventbased}. They are normally paired with other sensors such as image sensors and IMUs \cite{vidal2018ultimate, huang2024eventbased} for fast motion scenarios. The primary benefit of such an approach is to allow feature detection in the image space and feature tracking during the blind-time of the image sensor. To resolve feature associations, EVO \cite{EVO} combines two event sensors in a stereo configuration to enable feature matching across the two sensors. Similarly, EDS \cite{EDS} pairs a monocular camera with an event sensor for event-aided direct sparse odometry. Various learning based approaches also use event cameras for optical flow estimation \cite{bardow2016simultaneous,pan2020single,zhu2018ev}.

Despite its usefulness in robotics and vision-based navigation, \emph{event-only} SfO has not received a lot of attention in the literature. \citeauthor{Rebecq2017real-time}~\cite{Rebecq2017real-time} demonstrated visual-inertial odometry (VIO) with an event camera on a ``spinning leash'' sequence, which was obtained by spinning an event camera attached to a leash. However, the single result was evaluated only qualitatively~\cite[Fig.~7]{Rebecq2017real-time}. Moreover, being VIO, their method requires an onboard IMU.

\subsection{Event-based Feature Tracking}
\label{sec:tracking}

The first step in many VO pipelines is the feature detection and subsequent tracking across frames, which has proven to be difficult for the event-only case. Methods for corner detection \cite{eCDT,Mueggler17BMVC,li2019fa} have repurposed image feature detection methods to the event sensor. For feature tracking, methods either assume that known detections in the form of templates \cite{eKLT, HASTE} or use other sensors (such as images) for feature detection \cite{gehrig2018asynchronous}. Several methods have also explored learning based feature detection and tracking \cite{gehrig2018asynchronous,zhu2019unsupervised, chiberre2022longlived} mechanisms to compensate for the lack of image gradients and consistent pixel neighborhood across multiple view.

A serious challenge posed by a spinning object is the periodic self-occlusion unavoidably affecting features on the surface of the object. As we will show in Sec.~\ref{sec:results}, this leads to poorer quality tracks by existing methods. By reasoning over the spatio-temporal space of the event sensor to detect clusters of corner events that occur together and connect them over time to arrive at meaningful feature tracks, our method produces more accurate tracks.

\subsection{SfM for Single-axis Rotation}
The SfO problem explored in this work finds its roots in the earlier days of research into the SfM problem of ``single-axis rotation'' -- in which different methods explored reconstruction of objects placed on a turntable with a known rotational rate \cite{fitzgibbon1998automatic,fremont2004turntable}. These were further extended to include auto-calibration and recovery of camera poses along with the object reconstruction \cite{multipleUncalibrated2000}. The problem has made a reappearance recently in an ``in the wild'' context \cite{cheng2023accidental} where opportunistic turntable sequences allow object coverage due to the rotational motion using frame-based cameras. While their approach relies heavily on learning based techniques to form an implicit model of the object, our focus is more on geometric methods that require no prior training.
The method closest in application to ours is that of \citeauthor{Chen_2023}~\cite{Chen_2023}, who presented a dense reconstruction system using simulated event input, similar to a turntable sequence. The method generates dense object representations using a learning based framework, however, our focus is on estimating the rotational rate and axis as opposed to reconstructing a detailed model of the object. Similarly, the work of \citeauthor{baudron2020e3d}~\cite{baudron2020e3d} follows a shape from silhouette approach using an event sensor, mainly reconstructing the object shape.
Although \citeauthor{tweddle_2015_jfr_slam}~\cite{tweddle_2015_jfr_slam} formulate a similar problem in a similar setting, their work differs in that it focuses on modelling the motion of the object observed with stereo RGB cameras.
To our best knowledge, our work is the first to explore an event-only solution to recover the camera's orbital motion parameters over time jointly with a sparse 3D structure.

\subsection{Event Sensors for Space Applications}

In addition to high effective frame rate and high dynamic range, power consumption in milli-Watt levels makes event sensors attractive as an onboard sensor. Event sensors have already been applied to problems such as star tracking \cite{ng2022asynchronous,afshar2020event,chin2019star}, satellite material characterization \cite{doi:10.2514/1.A35015} as well as the general context of space situational awareness \cite{cohen2019event}.
In addition, there are several datasets for the application of event-based satellite pose estimation in space \cite{jawaid2022bridging,Park_2022,rathinam2023spades}.

\section{Event-based Structure-from-Orbit}
\label{sec:SfS}

\begin{figure}[t]
    \centering
    \includegraphics[width=0.49\textwidth]{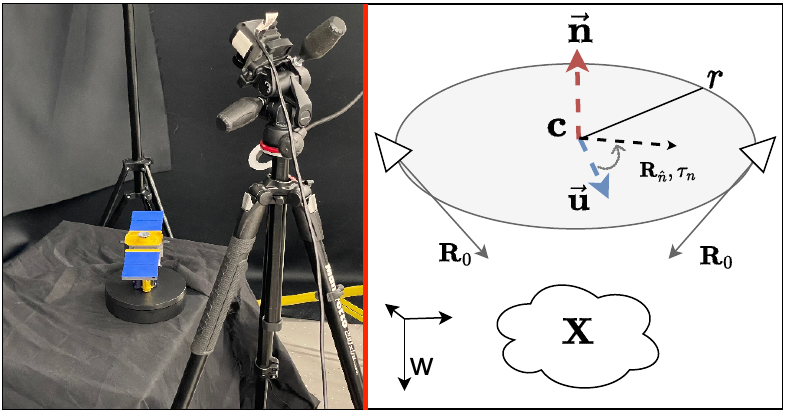}
    \caption{\textbf{L}: Problem setup. \textbf{R}: 
    Visualization of eSfO parameters in the orbit view of the problem. $w$ is an arbitrary world frame.}
    \label{fig:params}
\end{figure}

\begin{figure*}[!htb]\centering    \includegraphics[width=\textwidth]{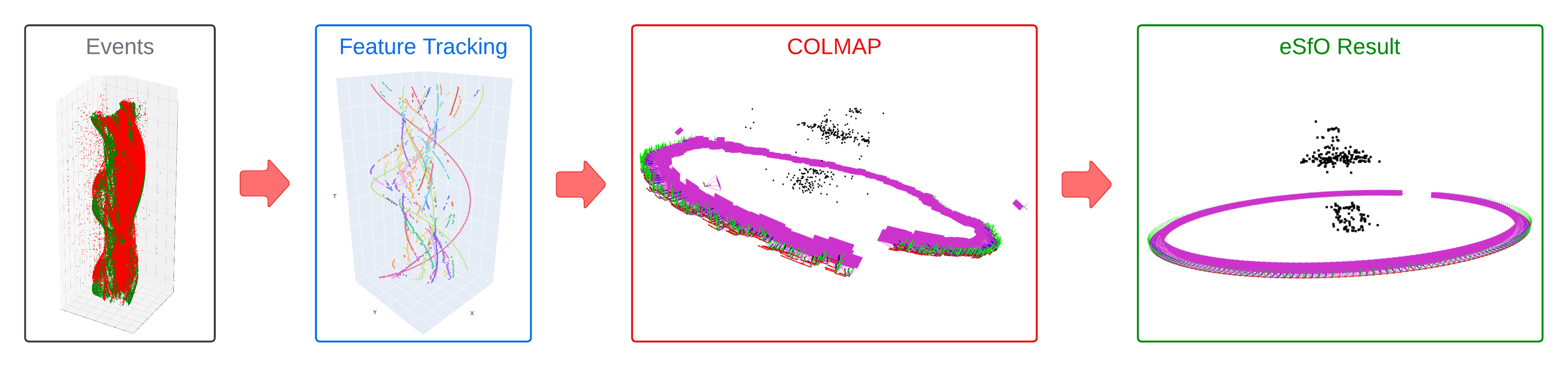}\\
\vspace{-1em}
    \caption{The proposed eSfO pipeline.}
    \label{fig:sfs-pipeline}
\end{figure*}

eSfO represents a special case of event-based SfM in which additional constraints are placed on the poses of the cameras. As alluded to in Sec.~\ref{sec:intro}, observing a spinning object with a static camera is equivalent to observing a static object with an orbiting camera; see supp.~material for the proof. 

The raw input stream from the event camera $\mathcal{E} = \{ e_i \}^{N_E}_{i=1}$, where each $e_i = (t_i, x_i, y_i, p_i)$, consists of the event pixel-locations ($x_i$,$y_i$), timestamp ($t_i$) and binary polarity ($p_i$). 
SfO recovers a sparse point cloud representation of the object $ \mbf{X} = \{ \mbf{x}_p^w \in {\mathbb{R}}^3 | p = 1 \dots N_P \}$ along with the following orbital parameters (see Fig. \ref{fig:params}):
\begin{itemize}
    \item $r \in \mathbb{R} $, the radius of the orbit.
    \item $f \in \mathbb{R} $, the rotational rate of the object (Hz).
    \item $\mathbf{R}_0 \in \mathcal{SO}(3)$, rotation with respect to the orbital plane that points the camera towards the object center
    \item $\Vec{\mbf{n}} \in \mathbb{R}^3$, unit vector representing the normal to the orbital plane pointing along the axis of rotation.
    \item $\Vec{\mbf{u}} \in \mathbb{R}^3$ lies in the plane, orthogonal to $\Vec{\mbf{n}}$.
    \item $\mathbf{c}^w \in \mathbb{R}^3$ which is the center of the 2D circle in 3D space.
    
\end{itemize}

The center of the camera lies on the circumference of the circle and its position in the world frame at time $\tau$ is
\begin{equation}
    \mathbf{t}^w(\tau; \mbf{\theta}) = r \cos{(2 \pi f \tau)} \Vec{\mathbf{u}} + r \sin{(2 \pi f \tau)}\Vec{\mathbf{v}} + \mathbf{c}^w,
\end{equation}
where 
$\Vec{\mathbf{v}} = \Vec{\mathbf{n}} 
\times 
\Vec{\mathbf{u}}$ lies within the plane and is mutually perpendicular to both 
$\Vec{\mathbf{n}}$ and $\Vec{\mathbf{u}}$. 
$\mbf{\theta}$ denotes the rest of the parameters.
The orientation of the camera can be decomposed into two rotations:
$\mathbf{R}_0$ which is a constant rotation with respect to the orbital plane that allows the camera to look at the object center, and $\mathbf{R}_{\Vec{\mbf{n}}, \tau}$ is the in-plane rotation around $\vec{\mbf{n}}$ at time  $\tau$ induced by the object's rotation.
This fully characterizes the orbital motion of the sensor. Compared to a general SfM problem involving $N$ cameras, where each camera has 6 DoF resulting in $6N$ parameters for the camera poses alone, the proposed formulation has a fixed number of motion parameters in this minimal representation (14 in total; see above) regardless of the number of cameras, due to the orbital constraint.
Additionally, there are $3N_{P}$ parameters required for the estimated landmarks.

The proposed eSfO pipeline  is depicted in Fig.~\ref{fig:sfs-pipeline}.
A front-end takes in the raw event stream and performs feature detection and tracking (Sec.~\ref{sec:tracking}). In the back-end, the set of tracks is used to recover an initialization  using a generalized SfM formulation (COLMAP \cite{COLMAP}). This initialization is upgraded using the eSfO optimization to conform to the orbital model of the problem (Sec.~\ref{sec:esfo_details}). 

\section{Event Feature Detection \& Tracking}\label{sec:tracking}

Event-only feature detection and tracking is an inherently difficult problem due to  the lack of consistent neighborhood structure and the asynchronous nature of the sensor. However, the temporal aspect of the event stream provides additional information to establish continuity and proximity over time. 
To address the problem of feature detection and tracking, we exploit densely clustered corner events in the spatio-temporal space.
Our approach, \textbf{Event Tracking by Clustering (ETC)}, hinges on two key observations: 
\textbf{(1)} The spatio-temporal density of corner events serves as a dependable metric for feature detection, as it is induced by the 3D structure of the scene; and 
\textbf{(2)} The spatio-temporal proximity of corner events
indicates whether these events originate from the same 3D point in space.

Based on these observations, we propose a feature detection and tracking mechanism taking advantage of the spatio-temporal nature of the event stream.

\subsection{Spatio-temporal Feature Detection}

Given $\mathcal{E}$, the first step in ETC is to detect ``corner'' events using the eFAST corner event detector~\cite{Mueggler17BMVC}.
eFAST leverages the Surface of Active Events (SAE) representation of the event stream.
An event $e_i$ is a corner when a certain number of neighboring pixels, located along a circular path around $(x_i,y_i)$, exhibit either consistently darker or brighter intensities in the SAE image; see~\cite{Mueggler17BMVC} for details. 

In our method, the detected corners are further filtered based on their local neighborhood density score, defined as
\begin{equation}
    D(e_i) := \frac{1}{\lambda} \sum _{e_q \in \mathcal{E}_i} \mathbb{I}(p_i = p_q),
\end{equation}
where $\mathcal{E}_i \subset \mathcal{E}$ is the set of events within distance $\lambda$ from $e_i$, $p_q$ is the polarity of a neighboring event $e_q = \{ t_q, x_q, y_q, p_q \}$, and $\mathbb{I}$ is the indicator function that returns 1 if its input condition is true and 0 otherwise.

Event cameras have different positive and negative polarity bias settings, leading to differing sensitivities for positive and negative polarity events~\cite{2023shining} -- and thus differing densities of corner events for the same brightness intensity edge.
We therefore compare the density score $D(e_i)$ of a corner $e_i$ with the mean density score $\mu_D$ of all detected corners of the same polarity as $e_i$, and remove $e_i$ if $D(e_i) < \mu_D$. This helps identify corner event clusters, which subsequently serve as candidate feature tracks.

\subsection{Spatio-temporal Feature Clustering}

We utilize our second observation to refine feature detections and track them by employing HDBSCAN \cite{mcinnes2017hdbscan}, a non-parametric, density-based clustering algorithm which identifies clusters by discovering regions of high data point density.
HDBSCAN is applied to the remaining corner events in the spatio-temporal space, disregarding polarity. This generates $N_H$ clusters of corner events, $C_i, i=\{1 \dots N_H\}$ that lay the foundation for our feature tracks.
Due to noise and the speed of motion in the scene, the tracks belonging to the same feature may be identified as disjoint clusters.
To join clusters that are spatio-temporally close to each other, 
we apply nearest neighbor head and tail matching within a spatio-temporal hemisphere (since time can only move forward).
The ``head'' represents the start of a cluster and ``tail'' represents the end. 
To connect clusters $C_i$ and $C_j$, 
the mean of the last and respectively first $N_\sigma$ events' spatio-temporal locations is computed and used as the descriptors for the clusters. Let these spatio-temporal descriptors be given by $D_\zeta = (t_\zeta, x_\zeta, y_\zeta) $ and $D_\eta = (t_\eta, x_\eta, y_\eta)$ for the tail and head of clusters $C_i$ and $C_j$ respectively. 
The cluster $C_i$ is connected to the cluster $C_j$ with the smallest euclidean distance between $D_\zeta$ and $D_\eta$ if it lies within the spatio-temporal hemisphere of radius $\phi$ defined by $D_\zeta$:
\begin{equation}
    (x_\eta - x_\zeta)^2 + (y_\eta - y_\zeta)^2 + (t_\eta - t_\zeta)^2 < \phi^2
\end{equation}
The two cluster are merged to form a single cluster, $C_i \gets C_i \cup C_j$, reducing the final number of clusters to $N_P$, corresponding to the number of tracks as well as the number of points in the reconstruction.

\subsection{Feature Track Extraction}
To extract feature tracks from the feature clusters, we divide the duration of the event stream, $T = t_{N_E} - t_{1}$, into $K = (T/\delta t)- 1 $ mutually disjoint temporal windows $\mathcal{W}_k, k = \{1 \dots K\}$ of duration $\delta t$ each 
such that the $k$-th window spans the time duration $
(k\delta t, (k+1)\delta t]$.
For each cluster $C_{p}, p = \{1 \dots N_P\}$, we
identify the events that fall within the time window $\mathcal{W}_k$: $\mathcal{E}_{p,k} = \{e_p | e_p \in C_p~\text{and}~ t_p \in \mathcal{W}_k\}$
and compute the mean pixel location within the window, giving us the position of the $p$-th feature track at time $t_k = k \delta t$:
\begin{equation}
     \mbf{f}^p_{t_k} = \frac{\sum_{ e_i \in \mathcal{E}{p,k}} (x_i,y_i)}
     {|\mathcal{E}_{p,k}|}
 \end{equation}

This approach has a few benefits: (1) the mean operation reduces noise of the corner events; and (2) we can choose the $\delta t$ to generate arbitrarily large number of camera views.
However, determining the appropriate window size $\delta t$ is application-specific, as it has an inverse relation with the tracking accuracy.

\section{Optimization for eSfO}\label{sec:esfo_details}

\subsection{Initialization}

The $N_P$ feature tracks generated using feature tracking provide the required information needed to run a general SfM pipeline. Each of the tracks correspond to a 3D point in the world frame $\mbf{x}^w_p$, who's projection at time $t_k$ is the observed feature $\mathbf{f}^p_{t_k}$. Each of the $K$ time windows correspond to a camera.
Using this information, we generate an initial set of cameras and a point cloud using COLMAP -- which minimizes the reprojection error for each 3D-2D correspondence $(\mbf{x}_p^w, \mbf{f}^p_{t_k})$ over the K cameras and $N_P$ world points:
\begin{equation}
    \sum_{k=1}^{K}\sum_{p=1}^{N_P} || \pi(\mathbf{KR}^{t_k}_w\mathbf{x}^w_p + \mathbf{K}\mbf{t}^{t_k}_w) - \mbf{f}^p_{t_k} ||
    \label{equ:reprojectionError}
\end{equation}
\noindent
where $t_k = k \delta t$, $\pi(.)$ is the pin-hole projection and $\mathbf{K}$ is the intrinsic calibration matrix. COLMAP provides an initial set of camera positions and 3D points, but since no constraints about the orbital trajectory were enforced in this general SfM pipeline, the camera trajectory deviates significantly from a circular path, see Fig.~\ref{fig:sfs-pipeline} (COLMAP). However, initial values for the SfO parameters can be computed from this intermediate solution. 

We compute the best fitting circle to the camera centers $\mathbf{t}^w_{t_k}, k = \{1 \dots K\}$ generated by COLMAP by estimating $\Vec{\mbf{n}}, \mbf{c}, \Vec{\mbf{u}}$ and the radius $r$.
To compute the normal vector $\Vec{\mbf{n}}$, we compute the mean of the camera centers $\mathbf{t_c} = \frac{1}{K} \sum_{k} \mathbf{t}^w_{t_k}$ and find the least squares fit for the plane using the $K \times 3$ mean normalized camera-center matrix $\mbf{T}$.
\begin{equation}
    \mbf{T} =
    \begin{bmatrix}
  \mathbf{t}^w_{t_1} - \mbf{t}_c, & \mathbf{t}^w_{t_2} - \mbf{t}_c, & \dots & \mathbf{t}^w_{t_k} - \mbf{t}_c 
\end{bmatrix}^{T}
\end{equation}
We solve $[\mbf{T}_{[:,1:2]}~\mathbf{1}_{K\times1}]\mbf{n} = \mbf{T}_{[:,3]}$ using least squares and normalize the resulting vector to obtain $\Vec{\mbf{n}}$. The notation $T_{[:,p:q]}$ selects all the rows and the $p$ to $q$ columns from the matrix $\mbf{T}$ and $\mbf{1}_{K\times1}$ is a $K\times 1$ matrix of ones. 

Camera centers are then projected to the computed plane to estimate the circle. The projection of the camera center $\mathbf{t}^w_{t_k}$ onto the plane is denoted as $\mbf{\Tilde{t}^w_{t_k}} = [Rod(\Vec{\mbf{n}}, \Vec{\mbf{z}})\mathbf{t}^w_{t_k}]_{[1:2]}$, where $Rod(\Vec{\mbf{a}}, \Vec{\mbf{b}})$ denotes the Rodriguez rotation matrix that rotates points about an axis $\Vec{\mbf{k}} = \Vec{\mbf{a}} \times \Vec{\mbf{b}}$ by an angle $\theta = \arccos{(\Vec{\mbf{a}}^T \Vec{\mbf{b}})}$.
The least square fit is found for the parameters of the circle $\mbf{\Theta}_c \in \mathbb{R}^3$ using the implicit equation of a circle. Using
\begin{equation}
    \mbf{\Tilde{T}} =
    \begin{bmatrix}
  \mathbf{\Tilde{t}}^w_{t_1} , & \mathbf{\Tilde{t}}^w_{t_2} , & \dots & \mathbf{\Tilde{t}}^w_{t_k} 
\end{bmatrix}^{T},
\end{equation}
we find the least square fit in the solution to the equation
\begin{equation}
    [\mbf{\Tilde{T}}~\mbf{1}_{K \times 1} ] \mbf{\Theta}_c = \begin{bmatrix}
  \mathbf{||\Tilde{t}}^w_{t_1}||^2 , & ||\mathbf{\Tilde{t}}^w_{t_2}||^2 , & \dots & ||\mathbf{\Tilde{t}}^w_{t_k}||^2 
\end{bmatrix}^{T}.
\end{equation}

The remaining orbit parameters are computed as:
\begin{align*}
    r &= \sqrt{\Theta_c(3) + (\frac{\Theta_c(1)}{2})^2 + (\frac{\Theta_c(2)^\star}{2})^2} \\
    \mbf{c}^w &= Rod(\Vec{\mbf{z}}, \Vec{\mbf{n}})\begin{pmatrix}
            \frac{\Theta_c(1)}{2}\\
            \frac{\Theta_c(2)}{2}\\
            0 
        \end{pmatrix} + \mbf{t}_c \\
    \mbf{u} &= \mbf{t}^w_{t_1} - \mbf{c}^w
\end{align*}
where the $(.)$ notation is used to access elements in the vector and
$\Vec{\mbf{z}} = (0, 0, 1)^T$.

Finally, for an estimate of the frequency,
we employ the Fourier Transform. 
We divide the duration of the event stream, $T = t_{N_E} - t_{1}$, into $F = (T/\delta t_f)- 1 $ mutually disjoint temporal windows $\mathcal{W}_f, f = \{1 \dots F\}$ (each of duration $\delta t_f$)
such that the $f$-th window spans the time duration $
(f\delta t_f, (f+1)\delta t_f]$. Within each window, we compute the mean value of the $x$ location of all the events. 
\begin{equation}
     \Tilde{x}(t_f) = 
     \frac{\sum_{e_i | t_i \in \mathcal{W}_f} x_i}
     {|\{e_i | t_i \in \mathcal{W}_f\}|}
\end{equation}
We then compute the Fourier Transform ($\Tilde{X}(f)$) of $\Tilde{x}(t_f)$.
\begin{equation}
    \Tilde{X}(f)\ \triangleq \ \int _{-\infty }^{\infty }{\Tilde{x}}(t_f)\ e^{-i2\pi ft}\ {\rm {d}}t
\end{equation}
The frequency with the highest power is used as the initial estimate of the rate of rotation.

\subsection{Solving for eSfO parameters}
\begin{figure}
    \centering
    \includegraphics[trim={8mm 1cm 4mm 0},clip,width=0.35\textwidth]{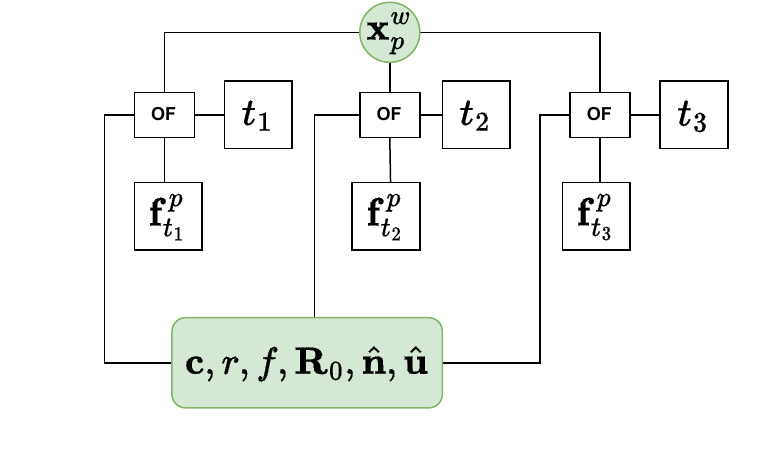}
    \caption{The formulation of the orbit factor (OF) illustrated for a single 3D point $\mathbf{x}_i$ and its corresponding feature positions for three different timestamps. Global parameters are highlighted as a group at the bottom of the figure. Estimated quantities are in green and inputs are in white squares.}
    \label{fig:orbitfactor}
\end{figure}

The eSfO parameterization consist of two sets of parameters: global parameters that capture the overall structure of the problem and local time-dependent observations of a 3D point in the world. The factor graph (See Fig.~\ref{fig:orbitfactor}) therefore includes dependencies between each observation induced by these global parameters. 

The formulation in its essence is a reprojection error minimization formulation similar to SfM, but differs in that the computation of the camera poses is instead governed by the orbit formulation. More concretely, we formulate the orbit-based camera transformation as
\begin{equation}
    \mbf{T}^{t_k}_w = \mbf{T}^{t_k}_{p} \mbf{T}_{o}^{p} \mbf{T}^{o}_{w},
\end{equation}
\noindent
which takes a point $\mbf{x}^w$ in a world frame and projects it to the camera coordinate frame at time $t_k$, where 
\begin{equation}
        \mbf{T}^{o}_{w} = \left[
        \begin{matrix}
            \mathbf{R}_{\Vec{\mbf{n}}}(2\pi f t_k ) & 0 \\
            0 & 1
        \end{matrix}
        \right]
        \left[
        \begin{matrix}
            Rod(\Vec{\mbf{n}}, \Vec{\mbf{z}}) & t^w(t_k ; \mbf{\theta}) \\
            0 & 1
        \end{matrix}
        \right] \\
\end{equation}
positions and orients the camera in the correct position along the circumference of the circle; 
\begin{equation}
    \mbf{T}^{p}_{o} = \left[
        \begin{matrix}
            & \Vec{\mbf{d}} && 0  \\
            & -\Vec{\mbf{y}} \times \Vec{\mbf{d}} && 0  \\
            & \Vec{\mbf{d}} \times (-\Vec{\mbf{y}} \times \Vec{\mbf{d}}) && 0  \\
            0 & 0 & 0 & 1
        \end{matrix}
        \right]
\end{equation}
rotates the camera to look at the center ($\mbf{c}^w$) of the orbit circle, with $\Vec{\mbf{d}} = \mbf{T}^{o}_{w}\mbf{c}^w/||\mbf{T}^{o}_{w}\mbf{c}^w||$, $\Vec{\mbf{y}} = [0, 1, 0]^T$; and 
\begin{equation}
        \mbf{T}^{t_k}_{p} = \left[
        \begin{matrix}
            \mbf{R}_{0} & 0  \\
            0 & 1
        \end{matrix}
        \right]
\end{equation}
\noindent 
rotates the camera from the orbital plane to toward the object. Points in the world frame $\mathbf{x}^w_p$ can then be projected to cameras where they are visible to minimize reprojection error, similar to (\ref{equ:reprojectionError}). This is used in a factor-graph method~\cite{GTSAM} to refine SfO parameters.

\section{Dataset}
\label{sec:dataset}
\begin{figure}[!htb]
\captionsetup{font=footnotesize}
\centering
\begin{subfigure}{0.235\textwidth}
  \centering
  \includegraphics[width=\textwidth]{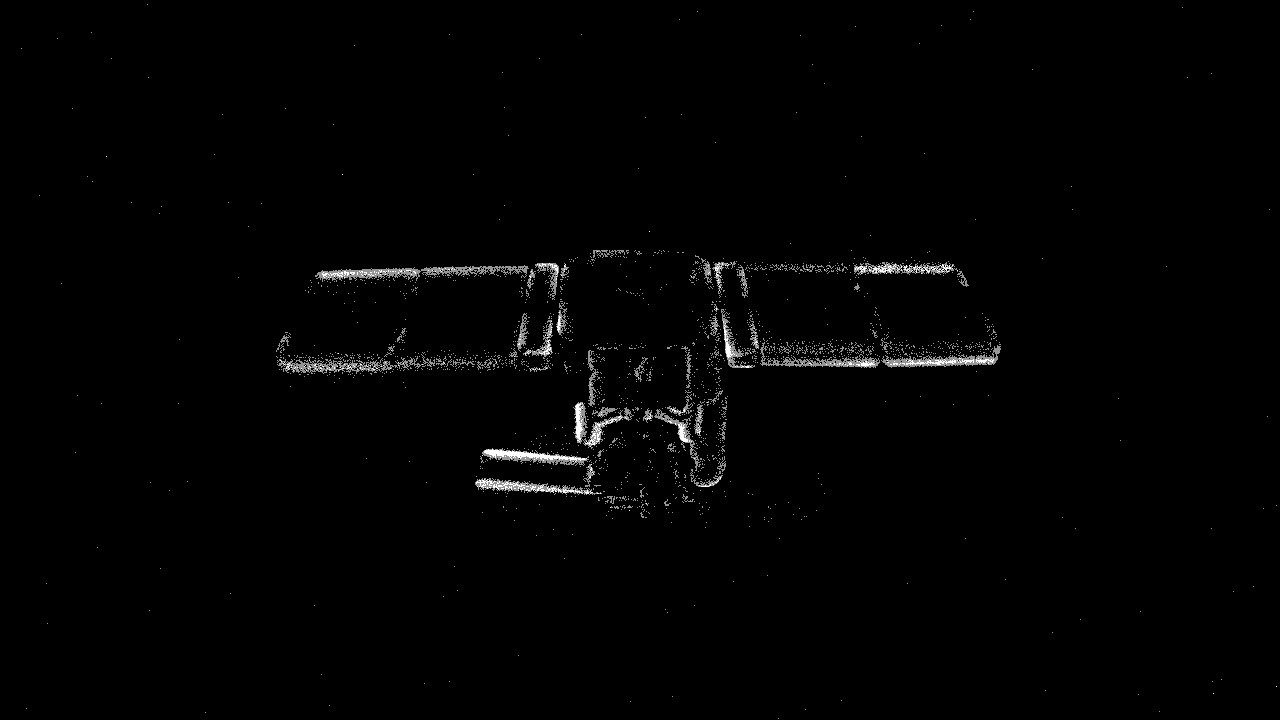}
\end{subfigure}
\begin{subfigure}{0.235\textwidth}
  \centering
  \includegraphics[width=\textwidth]{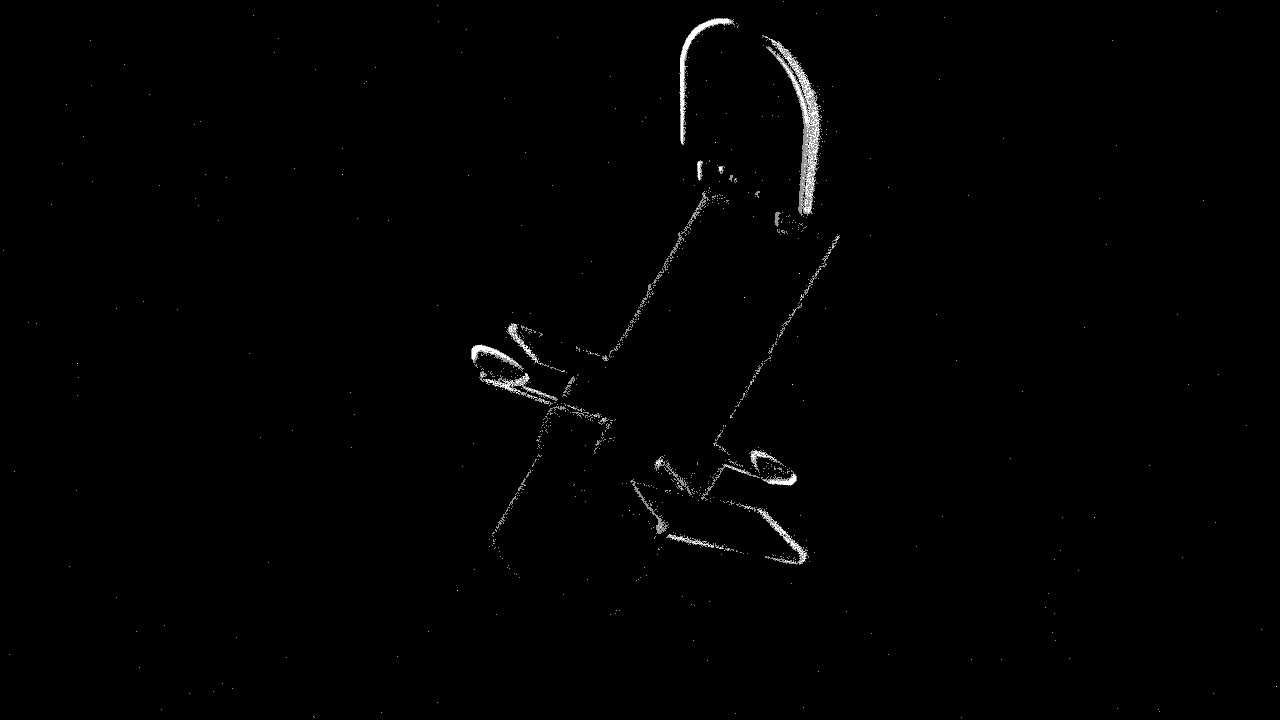}
\end{subfigure}
\begin{subfigure}{0.235\textwidth}
  \centering
  \includegraphics[width=\textwidth]{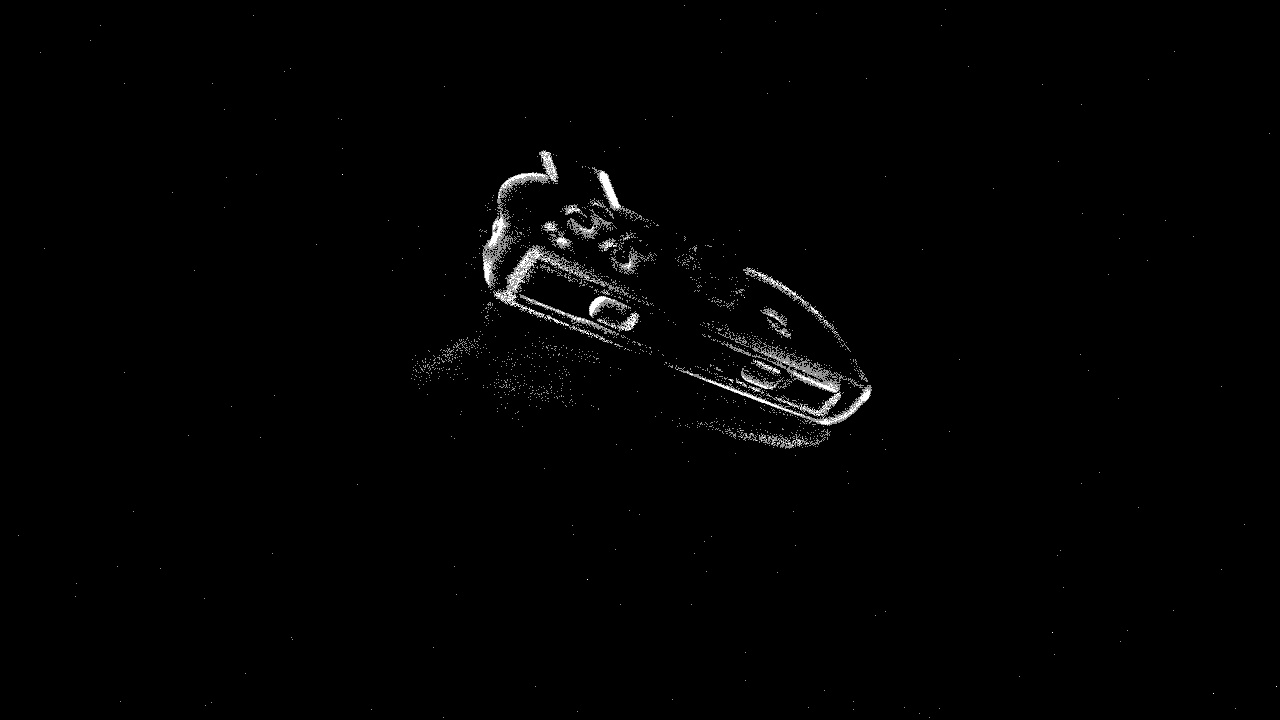}
\end{subfigure}
\begin{subfigure}{0.235\textwidth}
  \centering
  \includegraphics[width=\textwidth]{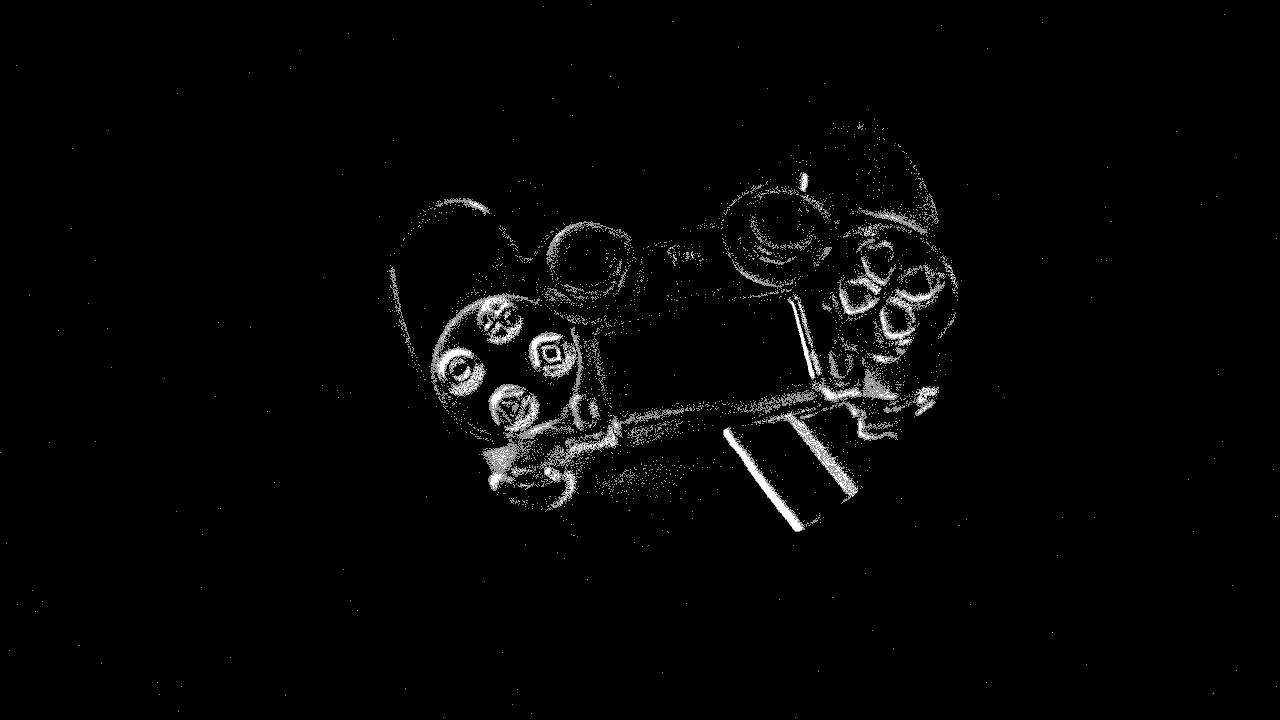}
\end{subfigure}
\caption[]{Example event frames from our dataset. \texttt{soho-sideon-fast} (top left), \texttt{hubble-diagonal-med} (top right), \texttt{switch-perpendicular-slow} (bottom left) and \texttt{dualshock-topdown-med} (bottom right).}
\label{fig:dataset_example}
\end{figure}

For performance evaluation,
we created a dataset of objects placed on a turntable and observed by a static event camera (Fig.~\ref{fig:params}).
Our \textbf{daTaset Of sPinning objectS with neuromorPhic vIsioN (TOPSPIN)}~\cite{elms_2024_10884694} consists of six objects under three rotational speeds and four perspectives (Tab.~\ref{tab:dataset_statistics}).
For each scene, one of the objects was placed on the turntable and rotated at a given speed, while the camera observed it from a given perspective. We exhaustively generated 72 scenes using the combinations in Tab.~\ref{tab:dataset_statistics}.
Several sample event frames are displayed in Fig.~\ref{fig:dataset_example}.

\begin{table}[h]
\centering
 \scalebox{0.7}{
\begin{tabular}{ |c|c|c| } 
 \hline
 Object & Turntable speed & Camera perspective \\ 
 \hline
 Hubble Satellite &  & \\ 
 SOHO Satellite & Fast & Top Down  \\ 
 TDRS Satellite & Medium & Side On \\ 
 PS4 Dualshock Controller & Slow & Perpendicular \\ 
 Nintendo Switch Controller & &Diagonal \\ 
 Inivation DAVIS346 Camera & & \\ 
 \hline
\end{tabular}
}
\caption{Taxonomy of settings for our event dataset.}
\vspace{-5mm}
\label{tab:dataset_statistics}
\end{table}

\section{Results}\label{sec:results}

We present experiment results to evaluate the efficacy of the two main contributions separately. For the front-end, we compared the performance of our feature tracker against state-of-the-art methods. For the back-end, we compared our method's sparse reconstruction against the objects' CAD model and compared the estimated frequency against the ground truth. We also provide qualitative results for the estimation of the axis of rotation of the objects.

\subsection{Hyperparameter settings}

The hyperparameters of the eSfO pipeline and their respective values are provided in Tab.~\ref{tab:parameter_settings}.
HDBSCAN has two hyperparameters $minPts$ and $\epsilon$ (see \cite{mcinnes2017hdbscan} for details).
\begin{table}[!h]
\centering
 \scalebox{0.7}{
\begin{tabular}{ |l|c| } 
 \hline
 Local event neighborhood distance ($\lambda$) & 7 \\ 
 \hline
 HDBSCAN minimum cluster size ($minPts$) & 10 \\ 
 \hline
 HDBSCAN cluster selection epsilon ($\epsilon$) & 5 \\ 
 \hline
 Track association hemisphre radius ($\phi$) & 30 \\ 
 \hline
 Cluster descriptor sample size ($N_\sigma$) & 5 \\ 
 \hline
 Feature track extraction window duration ($\delta t$) & 30ms \\ 
 \hline
 FFT sampling window duration ($\delta t_f$) & 20ms \\ 
 \hline
\end{tabular}
}
\caption{Parameter settings for eSfO.}
\vspace{-5mm}
\label{tab:parameter_settings}
\end{table}

\subsection{Front-end evaluation}

\paragraph{Event Camera Dataset}

\begin{figure}[]
\captionsetup{font=footnotesize}
\centering
    \includegraphics[width=0.5\textwidth]{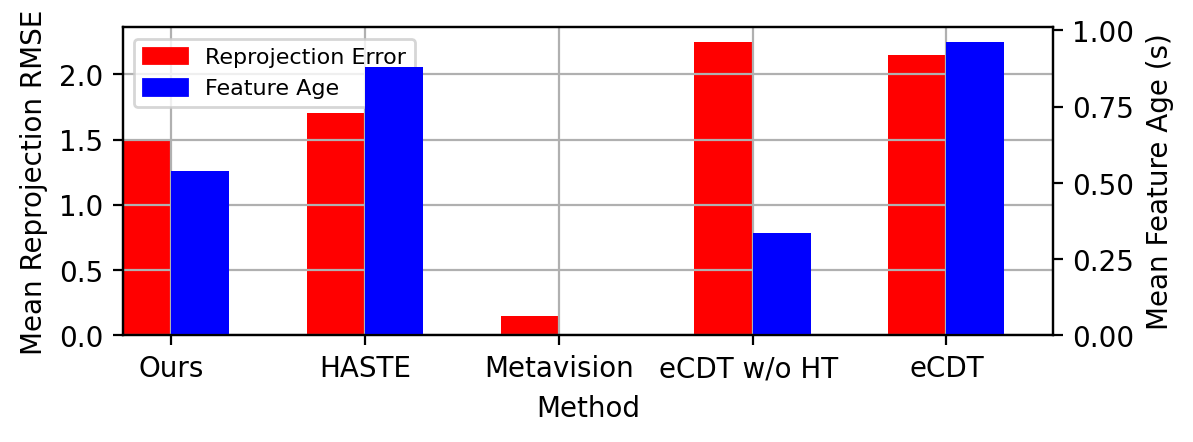}
\caption[]{Mean feature age and mean reprojection error computed across \texttt{shapes}, \texttt{poster} and \texttt{boxes}.}
\label{fig:feature-tracker-comparison}
\end{figure}

To evaluate ETC's performance against other state-of-the-art methods, we used the Event Camera Dataset~\cite{mueggler2017event} and the corresponding benchmarking strategy outlined in \cite{eCDT}: using ground truth camera poses, we triangulated the 3D points using the estimated feature tracks.
The estimated points were then projected to each ground truth camera and the reprojection error was computed against the tracked feature point.
Due to lack of a public implementation, results for eCDT are reported from their work \cite{eCDT}. eCDT without the HT Matching module is denoted as ``eCDT (w/o HT)''.
We initialized HASTE with the feature detections from our method, as it is only a tracking method.
Metavision \cite{chiberre2022longlived} used the pretrained weights from their original work.
Each dataset was evaluated with an error threshold of 3, 5 and 7 pixels.
Performance was then measured as the
Root Mean Squared Error (RMSE) of the projection error for each feature track and reported in Tab.~\ref{table:reproj_comparison}. 
Results for the feature age, the amount of time a feature is successfully tracked, are reported in Tab.~\ref{table:feature-age-comparison}. 
eCDT had the longest mean feature age, but suffered from a high mean reprojection error: their feature tracks were longer but less accurate.
Conversely, the Metavision feature tracker had extremely short feature tracks, which severely diminish its usefulness.
The ideal method would produce tracks which have both low reprojection error and longer feature age -- like the behavior seen for HASTE and ETC (Fig.~\ref{fig:feature-tracker-comparison}).

\paragraph{Event feature tracking for spinning objects}

\begin{figure}[!t]
\captionsetup{font=footnotesize}
\centering
\begin{subfigure}{0.23\textwidth}
  \centering
  \includegraphics[width=\textwidth]{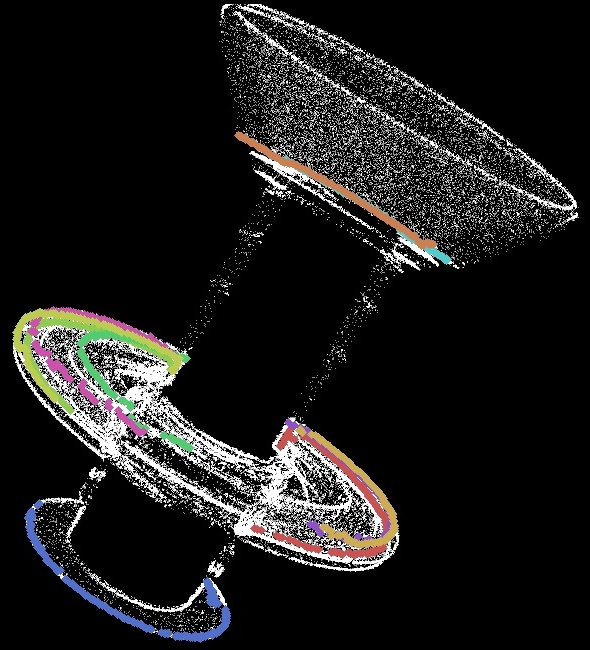}
  \caption{ETC}
\end{subfigure}
\begin{subfigure}{0.23\textwidth}
  \centering
  \includegraphics[width=\textwidth]{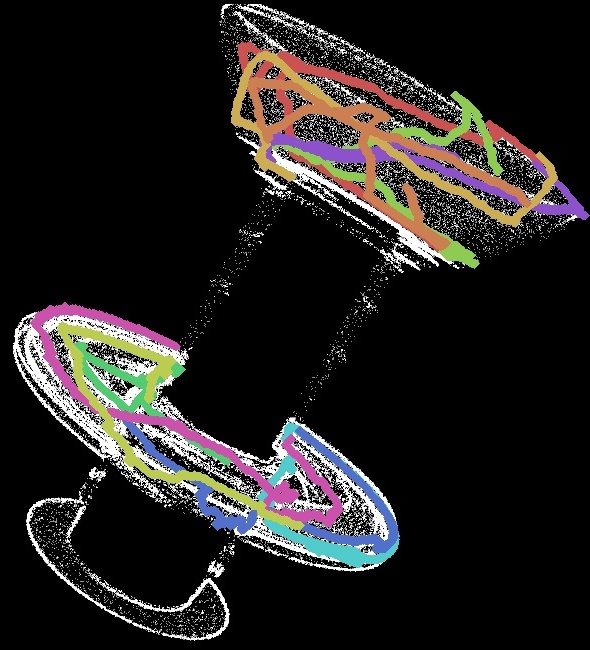}
  \caption{HASTE}
\end{subfigure}
\caption[]{10 longest feature tracks for both HASTE and ETC (our method) for the \texttt{hubble-diagonal-fast} scene.}
\label{fig:dataset-tracking-comparison}
\end{figure}

However, the underlying assumption of HASTE about the fixed (planar) tracking template during tracking is easily violated in the SfO setting, as observed when HASTE is evaluated on our dataset. Fig.~\ref{fig:dataset-tracking-comparison} shows feature tracks (colored) overlaid on accumulated corner events during a complete object revolution. Due to the object's rotation, the points on the object trace out an ellipse in the image \cite{Jiang_singleaxismotion}. HASTE tracks diverge from the elliptical arcs due to tracking failures. This is also reflected in the resulting COLMAP reconstruction, with a mean reprojection error of 348.48 pixels for the HASTE tracks.
In contrast, our method exhibited more robust tracking performance under the same conditions, with a mean COLMAP reprojection error of 2.39 pixels. See supplementary material for the full quantitative results.

\begin{figure}[!htb]
    \centering
\begin{subfigure}{0.23\textwidth}
  \centering
  \includegraphics[width=\textwidth]{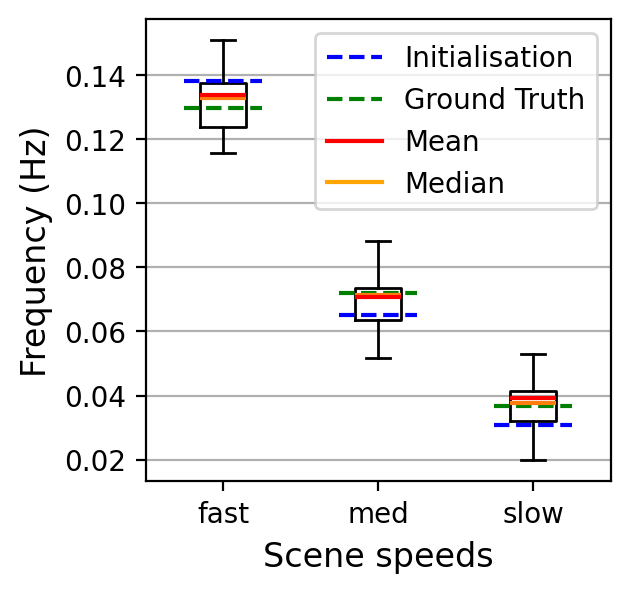}
    \caption{}
    \label{fig:scene-freq-dist}
\end{subfigure}
\begin{subfigure}{0.23\textwidth}
  \centering
  \includegraphics[width=\textwidth]{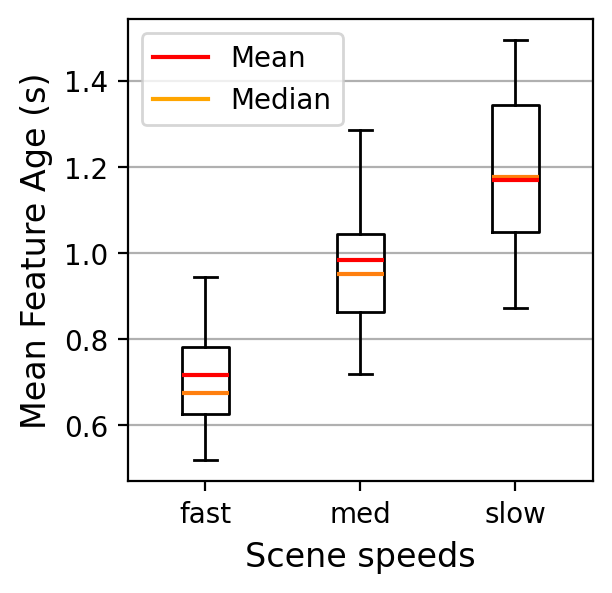}
    \caption{}
    \label{fig:scene-featage-dist}
\end{subfigure}
\caption{(a) Estimated frequency for the \texttt{fast}, \texttt{med} \& \texttt{slow} speeds across all scenes. (b) Distribution of mean feature age for the feature tracks from our method for all scenes; scenes are grouped by their speed (\texttt{fast}, \texttt{med} \& \texttt{slow}).}
\end{figure}

Note also that the feature age depends on the rotational velocity of the object (Fig.~\ref{fig:scene-featage-dist}). For faster motion, self-occlusion occurs faster than when observing slowly rotationing objects -- leading to a shorter mean feature age.

The results for the front-end feature detection and tracking show that our method is well-suited for the eSfO problem and compares favorably to other state-of-the-art methods on traditional benchmarks.


\begin{table}[!htb]
    \captionsetup{font=footnotesize}
	\centering
	  
	{
 \scalebox{0.7}{
	\begin{tabular}{l|l|ccc}
	\toprule \midrule
	&Method & 7 & 5 & 3 \\  \midrule
	\parbox[t]{2mm}{\multirow{5}{*}{\rotatebox[origin=c]{90}{\texttt{shapes}}}} &HASTE~\cite{HASTE} & 2.37 $\pm$ 1.87 & 1.73 $\pm$ 1.45 & 0.98 $\pm$ 0.85 \\
	&Metavision~\cite{chiberre2022longlived} & \textbf{0.17} $\pm$ 0.57 & \textbf{0.17} $\pm$ 0.57 & \textbf{0.17} $\pm$ 0.55 \\
	&eCDT(w/o HT)  & 3.10 $\pm$ 1.67 & 2.58 $\pm$ 1.25 & 1.76 $\pm$ 0.76 \\
	&eCDT & 2.94 $\pm$ 1.61 & 2.47 $\pm$ 2.32 & 1.78 $\pm$ 0.67  \\
	&ETC & \underline{1.70} $\pm$ 1.68 & \underline{1.26} $\pm$ 1.21 & \underline{0.88} $\pm$ 0.80  \\ \midrule 
	\parbox[t]{2mm}{\multirow{5}{*}{\rotatebox[origin=c]{90}{\texttt{poster}}}} & HASTE~\cite{HASTE} & \underline{2.30} $\pm$ 1.65 & 1.78 $\pm$ 1.24 & 1.17 $\pm$ 0.77 \\
	&Metavision~\cite{chiberre2022longlived} & \textbf{0.13} $\pm$ 0.45 & \textbf{0.13} $\pm$ 0.45 & \textbf{0.13} $\pm$ 0.44 \\
	&eCDT(w/o HT) & 2.95 $\pm$ 1.65 & 2.47 $\pm$ 1.22 & 1.75 $\pm$ 0.72 \\
	&eCDT & 2.69 $\pm$ 1.55 & 2.32 $\pm$ 1.17 & 1.71 $\pm$ 0.69  \\
	&ETC & 2.38 $\pm$ 2.09 & \underline{1.66} $\pm$ 1.50 & \underline{0.95} $\pm$ 0.92  \\ \midrule 
	\parbox[t]{2mm}{\multirow{5}{*}{\rotatebox[origin=c]{90}{\texttt{boxes}}}} &HASTE~\cite{HASTE} & 2.21 $\pm$ 1.78 & 1.68 $\pm$ 1.31 & 1.10 $\pm$ 0.79 \\
	&Metavision~\cite{chiberre2022longlived} & \textbf{0.18} $\pm$ 0.52 & \textbf{0.13} $\pm$ 0.45 & \textbf{0.13} $\pm$ 0.44 \\
	&eCDT(w/o HT) & 2.22 $\pm$ 1.54 & 1.94 $\pm$ 1.21 & 1.44 $\pm$ 0.74 \\
	&eCDT & \underline{2.12} $\pm$ 1.48 & 1.90 $\pm$ 1.20 & 1.42 $\pm$ 0.77  \\
	&ETC & 2.23 $\pm$ 2.07 & \underline{1.56} $\pm$ 1.50 & \underline{0.88} $\pm$ 0.91  \\ \midrule \bottomrule
	\end{tabular}
	}
 }
 \caption{RMSE (pixel) Reprojection errors. Reprojection errors above the threshold (7, 5, and 3) are considered outliers and removed.}
	\label{table:reproj_comparison}
\end{table}

\begin{table}[!t]
    \captionsetup{font=footnotesize}
	\centering
	
	{
 \scalebox{0.7}{
	\begin{tabular}{l|l|cccccc}
	
	\toprule \midrule
	&\multirow{2}[3]{*}{Method} & \multicolumn{2}{c}{7} & \multicolumn{2}{c}{5} & \multicolumn{2}{c}{3} \\  \cmidrule(lr){3-4} \cmidrule(lr){5-6} \cmidrule(lr){7-8} 
	&  & Mean & Med. & Mean & Med. & Mean & Med.   \\ \midrule
	\parbox[t]{2mm}{\multirow{5}{*}{\rotatebox[origin=c]{90}{\texttt{shapes}}}} &HASTE~\cite{HASTE} & \underline{0.646} & 0.200 & \underline{0.671} & \underline{0.270} & \underline{0.861} & \underline{0.430} \\
	&Metavision~\cite{chiberre2022longlived} & 0.007 & 0.005 & 0.007 & 0.005 & 0.007 & 0.005\\
	&eCDT(w/o HT) & 0.417 & \underline{0.240} & 0.427 & 0.235 & 0.446 & 0.227\\
	&eCDT & \textbf{1.224} & \textbf{0.550} & \textbf{1.309} & \textbf{0.585} & \textbf{1.518} & \textbf{0.628}  \\
	&ETC & 0.332 & 0.200 & 0.401 & 0.210 & 0.524 & 0.240  \\ \midrule
	\parbox[t]{2mm}{\multirow{5}{*}{\rotatebox[origin=c]{90}{\texttt{poster}}}} &HASTE~\cite{HASTE} & \textbf{1.699} & \textbf{1.107} & \textbf{1.186} & \textbf{0.829} & \underline{0.705} & \underline{0.481} \\
	&Metavision~\cite{chiberre2022longlived} & 0.005 & 0.005 & 0.005 & 0.005 & 0.005 & 0.005\\
	&eCDT(w/o HT) & 0.297 & 0.205 & 0.298 & 0.200 & 0.304 & 0.200\\
	&eCDT & \underline{0.795} & \underline{0.480} & \underline{0.814} & \underline{0.488} & \textbf{0.841} & \textbf{0.505}  \\
	&ETC & 0.499 & 0.180 & 0.497 & 0.180 & 0.641 & 0.190 \\ \midrule
	\parbox[t]{2mm}{\multirow{5}{*}{\rotatebox[origin=c]{90}{\texttt{boxes}}}} &HASTE~\cite{HASTE} & \textbf{0.904} & \textbf{0.685} & \textbf{0.721} & \textbf{0.554} & 0.524 & \underline{0.376} \\
	&Metavision~\cite{chiberre2022longlived} & 0.005 & 0.005 & 0.005 & 0.005 & 0.005 & 0.005\\
	&eCDT(w/o HT) & 0.275 & 0.200 & 0.277 & 0.200 & 0.283 & 0.200\\
	&eCDT & \underline{0.707} & \underline{0.380} & \underline{0.719} & \underline{0.385} & \underline{0.728} & \textbf{0.385}  \\
	&ETC & 0.525 & 0.180 & 0.702 & 0.200 & \textbf{0.738} & 0.200 \\ \midrule
	\bottomrule
	\end{tabular}
	}
 }
 \caption{Comparison of feature age (seconds). Reprojection errors above the threshold (7, 5, and 3) are omitted as outliers.}  
 
	\label{table:feature-age-comparison}
\end{table}

\subsection{Back-end}
In this section, we present quantitative and qualitative results for the various quantities estimated using eSfO and its efficacy at recovering the orbital structure. Again, see the supp.~material for the full quantitative results.

\paragraph{Frequency estimation}

eSfO aims to recover the rotation rate (frequency) of the object being observed. We compare the estimated frequency against the ground-truth obtained from the turntable. 
The results are presented in 
Fig.~\ref{fig:scene-freq-dist}, which depicts the average estimated frequency for the different speeds of the turntable. Observe that the initialization provided from the Fourier Transform is further refined by eSfO, bringing the estimate closer to the ground-truth. 

\paragraph{Is eSfO effective?}

To demonstrate the efficacy of eSfO at resolving the camera poses and structure, we first look at how the eSfO optimization behaves in the constrained orbit estimation problem. We report the average reprojection error for the eSfO reconstruction in Fig.~\ref{fig:sfo-error-dist}, and those for the COLMAP reconstruction in Fig.~\ref{fig:colmap-error-dist}. These plots show increased error for eSfO compared to COLMAP. However, eSfO imposes further constraints on camera poses to align them to a circular trajectory. This leads to more residual error after optimization, as not all constraints can be fully resolved. However, when comparing the generated trajectories from both (Fig.~\ref{fig:sfs-pipeline}), it is evident that eSfO recovers a better orbital trajectory compared to COLMAP. 

\begin{figure}[!h]
    \centering
    \includegraphics[width=0.5\textwidth]{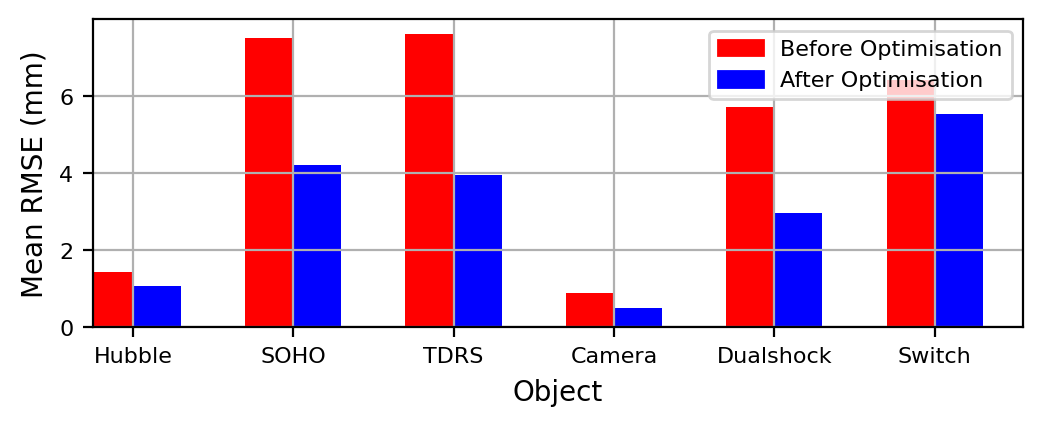}
\caption{Mean sparse point cloud to CAD model registration error before and after orbit optimization (all objects).}
\label{fig:model-reg-error}
\end{figure}

\begin{figure}[!htb]
    \centering
\begin{subfigure}{0.23\textwidth}
  \centering
  \includegraphics[width=\textwidth]{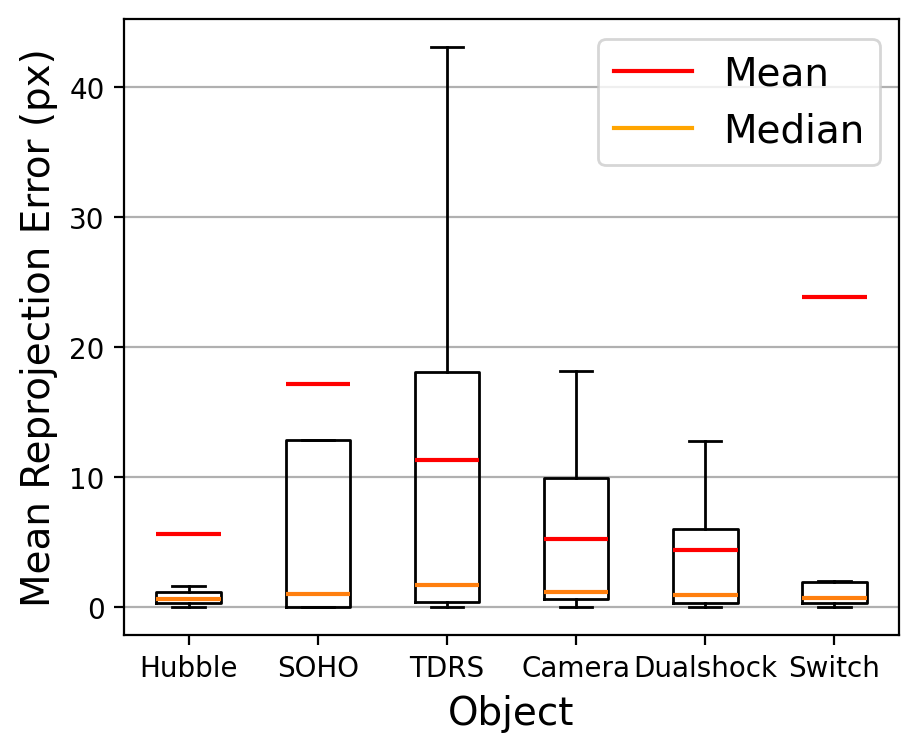}
    \caption{}
    \label{fig:sfo-error-dist}
\end{subfigure}
\begin{subfigure}{0.23\textwidth}
  \centering
  \includegraphics[width=\textwidth]{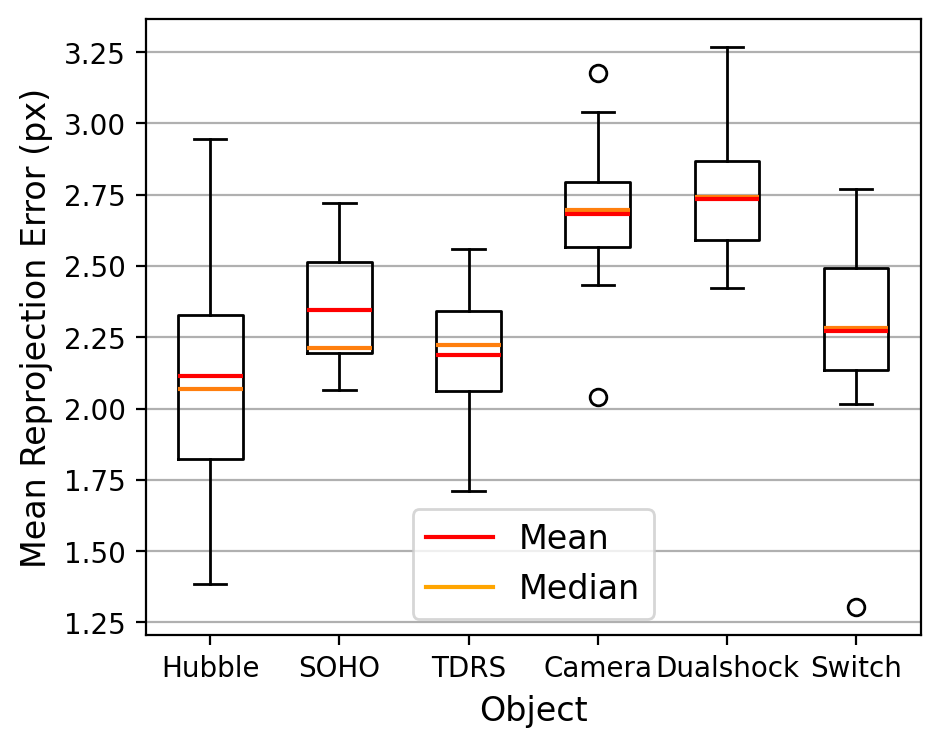}
\caption{}
\label{fig:colmap-error-dist}
\end{subfigure}
\caption{(a) Distribution of mean reprojection error after eSfO optimization. (b) Distribution of mean reprojection error of COLMAP reconstruction.}
\end{figure}

To investigate the effect of eSfO on point reconstruction, 
we aligned the point clouds with their respective 3D CAD models, initially using a manual alignment strategy and further refining it using an Iterative Closest Point (ICP) based optimization scheme. The results (Fig.~\ref{fig:model-reg-error}) show that the estimated point clouds are a highly accurate, albeit sparse, rendition of their respective 3D CAD model. Additional constructed trajectory and models are shown in the supplementary material.

\begin{figure}[!h]
\captionsetup{font=footnotesize}
\centering
\begin{subfigure}{0.235\textwidth}
  \centering
  \includegraphics[width=\textwidth]{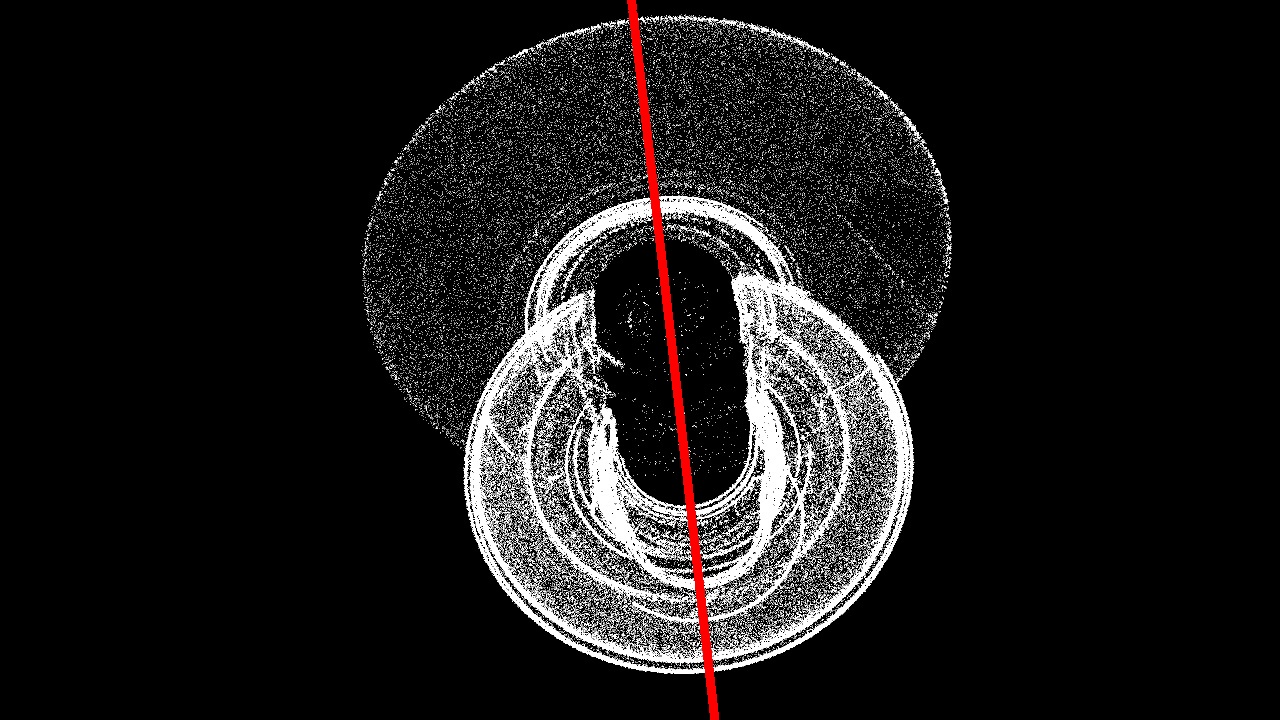}
\end{subfigure}
\begin{subfigure}{0.235\textwidth}
  \centering
  \includegraphics[width=\textwidth]{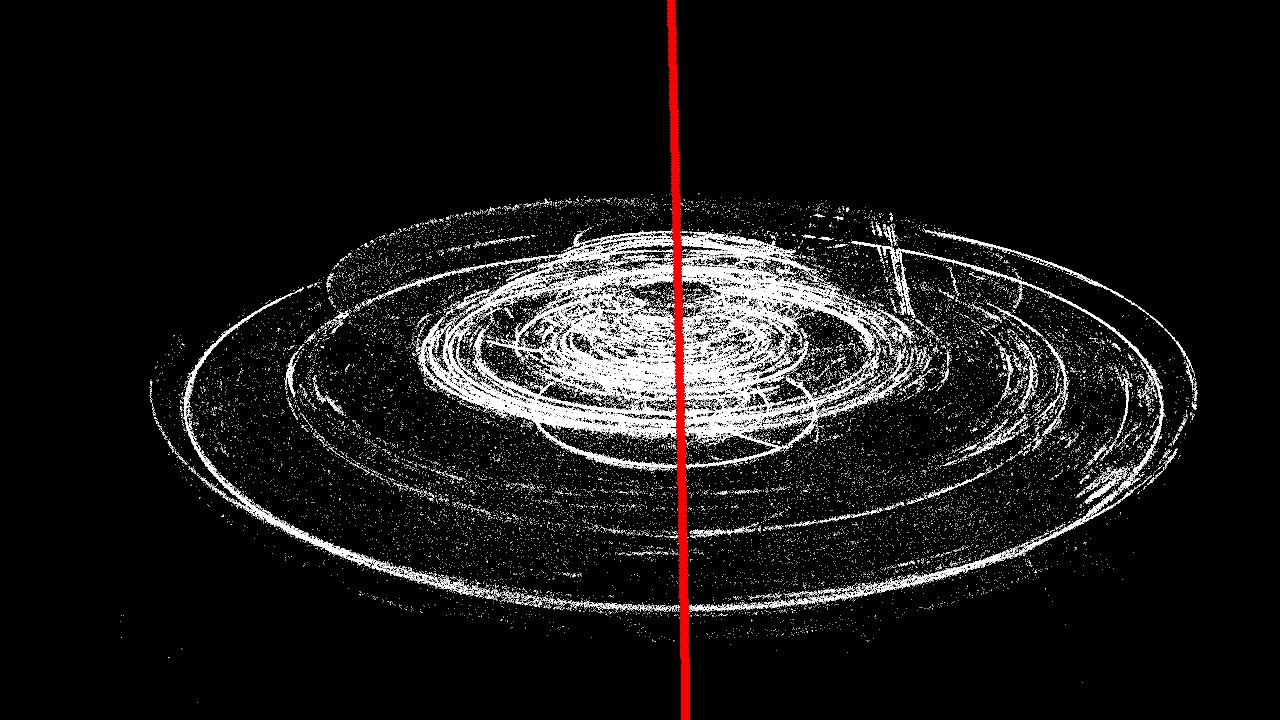}
\end{subfigure}
\begin{subfigure}{0.235\textwidth}
  \centering
  \includegraphics[width=\textwidth]{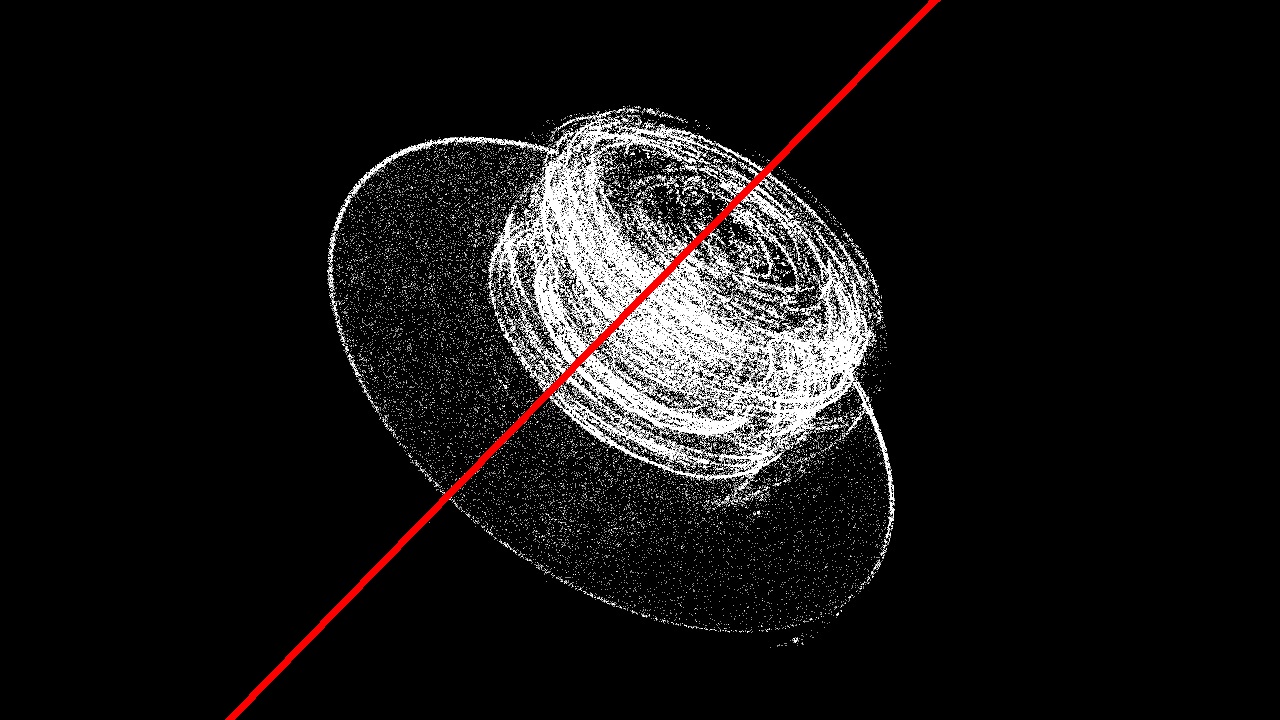}
\end{subfigure}
\begin{subfigure}{0.235\textwidth}
  \centering
  \includegraphics[width=\textwidth]{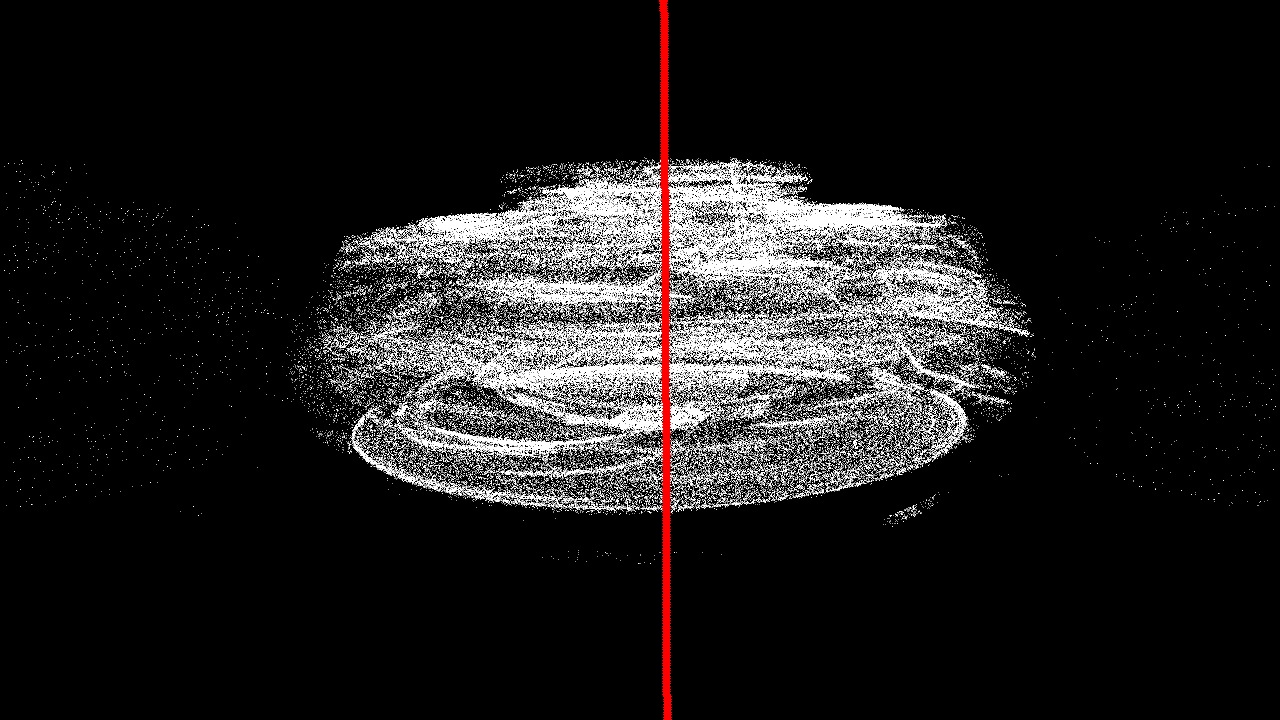}
\end{subfigure}
\caption[]{Estimated axis of rotation projected onto the accumulated event frame (red). \texttt{hubble-topdown-fast} (top left), \texttt{tdrs-sideon-med} (top right), \texttt{camera-diagonal-fast} (bottom left) and \texttt{dualshock-perpendicular-slow} (bottom right).}
\label{fig:estimated-axes}
\end{figure}

\paragraph{Axis of rotation estimation}
We present qualitative results for the projection of the axis of rotation into the image plane, referred to as the ``screw line''. This is the projection of the vector $\Vec{\mbf{n}}$ at the center $\mathbf{c}^w$ of the circle, onto the image plane. 
As seen in Fig. \ref{fig:estimated-axes}, the estimated screw line clearly marks out the axis of rotation of the object over the accumulated event frame -- suggesting that the plane normal and center of the circle were recovered correctly. 

\section{Conclusion}
In this work, we have presented a new reconstruction problem, \textit{eSfO}, which recovers a sparse representation of an object rotating about a fixed axis and observed by a static event camera. 
Through extensive experiments, we have demonstrated that the frequency, screw line, camera pose, and reliable sparse reconstruction can be recovered using the proposed pipeline.
The dataset has been released publicly~\cite{elms_2024_10884694}, and the code for eSfO can be found here:\\\url{https://github.com/0thane/eSfO}.

{
    \small
    \bibliographystyle{ieeenat_fullname}
    \bibliography{main}
}


\end{document}

%% file: preamble.tex
\usepackage[dvipsnames]{xcolor}
